\documentclass[letterpaper, 10 pt, conference]{ieeeconf}  

\IEEEoverridecommandlockouts                              

\usepackage{cite}
\usepackage{comment}
\usepackage{amsmath}
\usepackage{graphicx}
\usepackage{subcaption}
\usepackage{algorithm}
\usepackage[noend]{algpseudocode}
\usepackage{grffile}
\usepackage[colorinlistoftodos]{todonotes}
\usepackage[colorlinks=true, allcolors=blue]{hyperref}
\newcommand{\E}{\mathbf{E}}

\title{\LARGE \bf
Stochastic Variance Reduction for Policy Gradient Estimation}

\author{Tianbing Xu$^{1}$ and Qiang Liu$^{2}$ and Jian Peng$^{3}$
\thanks{$^{1}$Baidu Research, Sunnyvale, CA}%
\thanks{$^{2}$Department of Computer Science, Dartmouth College, New Hampshire}%
\thanks{$^{3}$Department of Computer Science, University of Illinois at Urbana-Champaign, Illinois}%
}

\begin{document}

\maketitle
\thispagestyle{empty}
\pagestyle{empty}

\begin{abstract}
Recent advances in policy gradient methods and deep learning have demonstrated their applicability for complex reinforcement learning problems. 
However, the variance of the performance gradient estimates obtained from the simulation is often excessive, 
leading to poor sample efficiency. 
In this paper, we apply the stochastic variance reduced gradient descent (SVRG) technique \cite{tong:svrg} to model-free policy gradient to significantly improve the sample efficiency.  
The SVRG estimation is incorporated into a trust-region Newton conjugate gradient framework 
\cite{nocedal:opt} for the policy optimization.
On several Mujoco tasks, our method achieves significantly 
better performance compared to the state-of-the-art model-free policy gradient methods in robotic continuous control such as trust region policy optimization (TRPO) \cite{schulman:trpo}. 
\end{abstract}


\section{Introduction}
Recently, policy gradient methods \cite{schulman:trpo, sutton:pg, williams:vpg, kakade:ng}
have achieved significant successes in 
challenging deep reinforcement learning problems.
Compared with value function based methods such as deep Q-learning \cite{mnih:dqn}, 
policy gradient methods directly optimize the density function 
to maximize the expected return, and works efficiently for large or continuous action spaces. 

Typical policy gradient (PG) is \emph{on-policy} 
in that they only make use of 
samples drawn from the current policy to update the policy parameters. 
In addition, the variance of gradient estimation in PG is often large, leading to poor or unstable convergence. 
This makes PG methods less sample efficient than Q-learning which are off-policy that can leverage all the samples from the whole history. 
This is a significant disadvantage given that drawing samples
in model-free RL settings is often expensive, requiring to run the simulator repeatedly. 

Trust region policy optimization (TRPO \cite{schulman:trpo}) is 
one of the state-of-the-art PG methods
that regularize the optimization using trust region techniques, 
and can achieve robust performance on a wide variety of challenging tasks. 
In this work, we propose to further improve the sample efficiency of TRPO by leveraging the stochastic 
variance reduced gradient descent (SVRG) technique \cite{tong:svrg} to improve the sample efficiency 
of the trust region updates.  
Our empirical results in Mujoco tasks show 
that SVRG technique significantly improves the efficiency of the algorithm, and our proposed practical method can achieve significantly better performance on some challenging tasks compared to TRPO. 
For the running results, please check the videos \footnote{https://drive.google.com/drive/folders/14E9yRCK8Wzbf3SqMggXa1SioJLz3gFQx}.


\section{Background}
\subsection {Reinforcement Learning}
Consider an agent operating in an uncertain environment. 
At each time step $t$, the agent takes an action $a_t$  based on 
its observation of the environment state $s_t$,
 subsequently observes a new state $s_{t+1}$, and receives a scalar reward $r_t$.   
The goal of reinforcement learning is to find an optimal policy  $\pi(a_t |s_t)$, 
for choosing action $a_t$ given an observation $s_t$, 
following a distribution of $a_t$ conditional on $s_t$, 
to maximize the expected return:
\begin{equation}
\label{eq:loss}
L(\pi) = \mathbf{E}_{\pi} \big[\sum_{t=0}^{\infty} \gamma^t r_t(s_t, a_t)\big], 
\end{equation}
where $\gamma \in (0,1)$ is a discount factor, and $\mathbf E_{\pi}$ does the expectation under the policy $\pi(a_t|s_t)$ and the (unknown) stochastic environmental dynamics. 
For notational simplicity, we provide definitions of three commonly used value functions as following.

Given a policy $\pi$, the state value function is defined as the expected return from state $s_t$ according to the policy $\pi$ in the environment:  
\begin{align*}
V^{\pi}(s_t) = \mathbf{E}_{a_t, s_{t+1}, ...} \big[ \sum_{t'=t}^{\infty} \gamma^{t'-t} r_{t'} (s_{t'}, a_{t'})  \big]. 
\end{align*}
The state-action function (i.e., the Q function) is defined as the expected return 
when taking action $a_t$ from state $s_t$: 
\begin{align*}
Q^{\pi}(s_t, a_t) =  \mathbf{E}_{s_{t+1}, a_{t+1}...} \big[ \sum_{t'=t}^{\infty} \gamma^{t'-t} r_{t'} (s_{t'}, a_{t'})  \big]. 
\end{align*}
The advantage function is defined as the state-action function subtracted by the state value function,
\begin{align*}
A^{\pi} (s_t, a_t) = Q^{\pi}(s_t, a_t) - V^{\pi}(s_t). 
\end{align*}
It is worth noting that the advantage function measures the relative advantage value of action $a_t$ compared with 
the average return when following the policy $\pi$ from state $s_t$. 

When the cardinality of the set of all states and actions is small, policies can be readily parameterized as conditional probability tables (CPTs). However, for most real-world tasks, such as video games and robotic manipulation, it will be intractable to fully parameterized policies using CPTs. Functional approximations, such as neural-network, have been shown applicable on various applications. 
For example, Gaussian MLP is used in robotics control tasks such as Mujoco\cite{todorov:mujoco}.

\subsection{Policy Gradient Methods}
Policy gradient methods have been proposed to optimize a parameterized policy $\pi$ with respect to the return $L(\pi)$ via gradient descent.

Assume $w$ is the parameter of a policy $\pi_w(a_t | s_t)$. For simplicity, we use $L(w)$ and $L(\pi_w)$ interchangeably. According to the policy gradient theorem \cite{sutton:pg}, the gradient of the return $L(w)$ can be derived as following, 
\begin{align}
\label{eq:pg}
\nabla_w L(w) = \E_{\pi_w} [\sum_{t=0}^\infty \nabla_w \log \pi(a_t | s_t) Q^\pi(s_t, a_t)]. 
\end{align}
Since $Q^\pi$ is difficult to compute, in practice we draw $M$ roll-out trajectories $\tau^i = \{s_t^i, a_t^i, r_t^i\}_{t=1}^T$ from the current policy $\pi_w$, and estimate the gradient by 
\begin{align}\label{equ:wl}
\nabla_w L(w) \approx 
\frac{1}{M}\sum_{i=1}^M \sum_{t=1}^T \nabla_w \log \pi_w(a_t^i | s_t^i) R^i_t, 
\end{align}
where the future $R_i^t = \sum_{t' = t}^T \gamma^{t'-t} r_{t'}^i$ is an unbiased estimator of $Q^\pi(a_t^i, s_t^i)$. 
To reduce the variance of the gradient estimation, 
we can subtract a state-dependent baseline function from the future return without introducing any estimation bias. 
For example, \textbf{REINFORCE} \cite{williams:vpg} 
replaces $R_t^i$ with $R^i_t - b$ where $b$ is a constant baseline, 
and advantageous actor critic (A2C) uses $R^i_t - \hat V^\pi(s_t^i)$, where $\hat V^\pi$ is an estimator of the value function $V^\pi$. 

\subsection{Trust Region Policy Optimization}
The efficiency and convergence of policy gradient can be significantly improved by 
considering the information geometry of the space of policy distributions. A notable example is trust region policy optimization \cite{schulman:trpo}. Instead of simply performing gradient descent of policy parameter, TRPO enforces a KL divergence constraint on a local metric of policy distributions as opposed to the Euclidean metric in parameter space. In particular, the parameter update in TRPO can be formulated as a constrained optimization as following.

\begin{equation}\label{equ:trpo}
\begin{aligned}
w^{\ell+1} = & \arg\max_w
& & U_\ell(w) = \mathbf{E}_{\pi_{w^{\ell}}} \big[ \frac{\pi_w(a|s)}{\pi_{w^{\ell}}(a|s)} A^{\pi_{w^{\ell}}}(s,a) \big] \\
& \text{s.t.}
& & \overline{D}_{KL}(\pi_{w^{\ell}}, \pi_w) \leq \delta
\end{aligned}
\end{equation}
where $U_{{\ell}}(w)$ is a "surrogate objective" that
serves as a local approximation to the expected return $L(w)$. 
This surrogate objective matches the exact objective when $w = w^\ell$, i.e., 
\begin{align*}
U_\ell(w) = L(w), && \nabla_w U_\ell(w) = \nabla_w L(w), 
\end{align*}
while providing an approximation of $L(w)$ when $w \approx w^{\ell}$.  
In Eq \eqref{equ:trpo}, the $\overline{D}_{KL}$ is the expected KL divergence of states observed with visitation probability induced by $\pi_{w^\ell}$, defined as 
\begin{align*}
\overline{D}_{KL}(\pi_{w^\ell}, \pi_w) = \mathbf{E}_{s_t \sim \pi_{ w^\ell}} \big[ D_{KL}(\pi_{w^\ell}(\cdot |s_t), \pi_w(\cdot  | s_t))\big], 
\end{align*}
and $\delta$ is a tunable parameter to upper bound the difference between the consecutive policies to ensure locality of the approximation and the stability of policy updates. 
With this constraint, $\pi_w$ would be not too different from the previous policy $\pi_{w^\ell}$.
When $\delta$ approaches zero, TRPO is reduced to the Natural Policy Gradient \cite{kakade:ng}, 
\begin{equation}\label{equ:fisher}
w^{\ell+1} \gets w^{\ell} + \eta  H^{-1}(w^\ell) \nabla_w L(w^{\ell}), 
\end{equation}
where $H(w^\ell)$ is the Fisher information matrix of policy $\pi_{w^\ell}$, $H(w^\ell) = \mathbf{E}_{a,s \sim \pi_{ w^\ell}} [\nabla_{w^\ell} \log \pi_{w^\ell}(a | s) \cdot \nabla_{w^\ell} \log \pi_{w^\ell}(a | s)^T].$ 
In practice, 
TRPO is implemented by approximately solving \eqref{equ:trpo} 
using natural gradient update \eqref{equ:fisher}, with a line search on the step size $\eta$ 
to meet the expected KL divergence constraint.

\section{Trust Region Stochastic Variance Reduction Policy Optimization}

In practice, each iteration of TRPO rollouts a set of samples $\{(s_t, a_t, r_t)\}_{t=1}^N$ according to the policy $\pi_{w^\ell}$, 
and approximates the surrogate function with 
\begin{align}\label{equ:hatU}
\hat U_\ell(w) = \sum_{t=1}^N  \rho_t \hat A^{\pi_{w^{\ell}}}(s_t,a_t)
\end{align}
where $\rho_t = {\pi_w(a_t|s_t)}/{\pi_{w^{\ell}}(a_t|s_t)}$ denotes the density ratio between $\pi_w$ and $\pi_{w^\ell}$, 
and $ \hat A^{\pi_{w^{\ell}}}(s_t,a_t) = R_t - \hat{V}^{\pi_{w^\ell}}(s_t)$ is an estimation of the advantage function. 
So the gradient can be computed by 
\begin{align}\label{equ:dhatU}
\nabla  \hat U_\ell(w)  
= \frac{1}{N} \sum_{t=1}^N \nabla U^t_{\ell} (w), 
\end{align}
where each $\nabla_w U^t_{\ell} (w) =  \rho_t \nabla_w \log \pi_w(a_t | s_t) \hat A^{\pi_{w^{\ell}}}(s_t,a_t)$ is the gradient contributed by the $t$-th sample 
 $(s_t, a_t, r_t)$. 

 

\subsection{Stochastic Variance Reduced Gradient}

Stochastic variance reduced gradient (SVRG)   is variant of an accelerated stochastic gradient descent method that 
achieves faster convergence using variance reduction by a novel control variate \cite{tong:svrg}.  
 Here we give a brief review on SVRG in this section. 

Assume we are interested in optimizing our objective function $\hat U_\ell(w) = \sum_{t=1}^N U^t_\ell(w)/N$ where $N$ is large. 
Typical stochastic gradient estimates the gradient by sub-sampling:
\begin{align}
\label{eq:minipg}
\hat\nabla \hat U_\ell(w) = \frac{1}{m}\sum_{t\in \mathcal I} \nabla U^t_\ell(w)
\end{align}
where $\mathcal I  \subset 1:N$ is a mini-batch of size $m$. A choice of a small batch size $m$ improves the computation efficiency but causes larger variance. 
SVRG allows more accurate estimation of the gradient with small $m$. 
The idea is to maintain a snapshot parameter $\tilde w$
that is close to the current parameter $w$ (i.e., let $\tilde w$ be the parameter in the past few iteration),  
 whose accurate gradient (i.e., $\nabla \hat U_\ell(\tilde w) = \sum_{t=1}^N \nabla U^t_\ell(\tilde w)/N)$ is pre-calculated, 
and use it as an anchor point for estimating $\nabla \hat U_\ell(w)$. 
Specifically, SVRG estimates $\nabla \hat U_\ell(w)$ by 
\begin{align}
\label{eq:spg}
\hat {\nabla}_\mathrm{sv} U_\ell(w) =  \nabla \hat U_\ell(\tilde w)  + \frac{1}{m} \sum_{t\in \mathcal I} {\left(\nabla U^t_\ell(w) - \nabla {U^t_\ell(\tilde{w})}\right)}. 
\end{align}
The key is that we use the mini-batch to estimate the difference 
$\nabla U^t_\ell(w) - \nabla {U^t_\ell(\tilde{w})}$, which can be small when $w$ is close to $\tilde w$. 
Specifically, SVRG gives an unbiased estimation of the exact gradient $\nabla \hat U_\ell(w)$,
with variance $\frac{1}{m}\mathrm{var}_{t\in\mathcal{I}}(\nabla U^t_\ell(w) - \nabla {U^t_\ell(\tilde{w})})$, which is small when $w \approx \tilde w$. 
In contrast, the naive stochastic gradient has a variance $\frac{1}{m} \mathrm{var}_{t\in\mathcal{I}}(\nabla U^t_\ell(w))$ which can be much larger. 

\begin{algorithm*}
\caption{Trust Region Stochastic Variance Reduction for Policy Optimization}\label{svrg}
\begin{algorithmic}[1]
\Procedure{Stochastic Variance Reduction Policy Optimization}{$N$, $L$, $J$, $m$, $\nu$}
\State \textbf{Inputs:}
\State{$N$: number of transitions ($s_t$, $a_t$, $r_t$, $s_{t+1}$) in each iteration}
\State{$L$: number of outer loop iterations (epochs)}
\State{$J$: maximum number of mini-batches}
\State{$m$: mini-batch size}
\State{$\nu$: sub-sampling ratio for Fisher information matrix}

\State \textbf{Initialization:}
\State Initialize $\tilde{w}^0$ randomly
\For{$\ell = 1$ to $L$}
\Comment{each epoch}
\State{Generate $N$ transitions by executing the current policy $\pi_{\tilde{w}^\ell}$. 
Calculate the baseline gradient:}
\vspace{.3\baselineskip}
\State {$\tilde{g}^\ell = \frac{1}{N} \sum_{t=1}^N {\nabla U^t_\ell(\tilde{w}^\ell)}$}. 
\vspace{.3\baselineskip}
\State {Initialize the inner loop: $w_0^\ell = \tilde{w}^\ell$.}
\For{$j = 1$ to $J$}
\Comment{mini-batch}
\State {Draw a mini-batch $I_j$ (with size $m$) uniformly random from $1:N$}
\State{Calculate the SVRG estimation $\hat{g}(w)$:}
\begin{align}
\hat\nabla_{\mathrm{sv}} U_\ell(w) = 
 \tilde{g}^\ell + \frac{1}{m}\sum_{t\in I_j} (\nabla U^t_\ell(w) - \nabla U^t_\ell(\tilde w^\ell)). 
\end{align}
\State{Update policy parameter with mini-batch $I_j$:}
\State{
\begin{equation}
\label{eq:newton}
w_{j +1}^\ell \gets w_j^\ell + \eta_j \hat{H}^{-1}(w_j^\ell) \hat\nabla_{\mathrm{sv}} U_\ell(w_{j}^\ell). 
\end{equation}
}
\State{where Fisher information matrix $\hat{H}(w_j^\ell)$ is estimated with $N \nu$ samples.
We use conjugate gradient to compute $\hat{H}^{-1}(w_j^\ell) \hat{g}(w^\ell_j) $, 
and use backtracking line search to calculate step size $\eta_j$.}
\EndFor
\State{$\tilde{w}^{\ell+1} \gets w_{J}^\ell$} \label{alg:snapshot}
\Comment{update policy}
\EndFor
\State \textbf{Output:}
$w \gets \tilde{w}^{L}$
\EndProcedure
\end{algorithmic}
\end{algorithm*}

\subsection{Stochastic Variance Reduced Policy Optimization}
In order to improve the efficiency of TRPO, we apply SVRG to optimize the surrogate $U_\ell(w)$ in \eqref{equ:trpo}, 
with $w^\ell$ as the anchor point. 
Our main algorithm, Stochastic Variance Reduced Policy Optimization (SVRPO), 
is summarized in Algorithm~\ref{svrg}. 
Our algorithm is a double loop method, 
where in the $\ell$-th outer loop 
we draw $N$ transition samples by performing the current anchor policy $\pi_{w^\ell}$, 
and construct an approximate surrogate function $\hat U_
\ell(w)$ in Eq \eqref{equ:hatU}.
In the inner loop iterations, 
we optimize $\hat U_\ell(w)$  using SVRG, initializing from the anchor policy $w^\ell$. 
Similar to TRPO, the KL constraint $\bar D_{KL}(\pi_{w^\ell}, ~ \pi_w) \leq \delta$ is addressed by proper selection of the step size. 
In particular, we use back tracking line search to tune the step size $\eta$ by iteratively halving the step size until the expected return utility function is improved within maximum number of iterations, and further the initial step size is upper-bounded by the policy distance $\delta$ under the KL constraint \cite{schulman:trpo}.
 
For practical efficiency, we use a sub-sampled trust region Newton conjugate gradient method with SVRG, similar to \cite{nocedal:hessian,nocedal:subsample,erdogdu:svrg,zhang:vrp}, that is, 
$$
w^\ell_{j+1} \gets w^\ell_j + \eta_j \hat H^{-1}(w^\ell_j) \hat \nabla_{\mathrm{sv}} \hat U_\ell(w^\ell_j), 
$$
where $w^\ell_j$ is the parameter at the $j$-th inner loop of the $\ell$ iteration, 
and $\hat H(w^{\ell}_j)$ is an approximation of the Fisher information of $\pi_{w^\ell_j}$,
defined as the empirical covariance matrix of $\nabla_{w^\ell} \log \pi_{w^\ell}(a | s)$ based
on a mini-batch (of size $N\nu$) of samples. 
In addition, we use conjugate gradient to calculate the multiplication of matrix inversion and SVRG gradient 
vector for computational efficiency.   

\section{Experimental Results}
\label{sec:exp}
We compare our method to TRPO, the state-of-the-art policy optimization algorithm in various 
robotics control tasks in Mujuco,
and demonstrate the superiority of our proposed stochastic variance reduction policy 
optimization method in term of sample efficiency and performance. Furthermore, we 
investigate the performance and convergence acceleration effects by variance reduction and 
Fisher information. 

\subsection{Comparison with TRPO in Mujoco Tasks} 
\label{sec:mujoco}
TRPO has been a strong baseline and has been shown to perform robustly on a wide variety of tasks, especially in Mujoco robotics controls \cite{duan:benchmark}. 
We find that our method significantly improves the sample 
efficiency and achieves significantly better performance in many tasks such as Swimmer, Walker, Ant and Hopper, and gets similar return in the remaining tasks. 

\begin{figure*}
\centering
\begin{subfigure}[b]{0.3\textwidth}
\includegraphics[width=\textwidth]{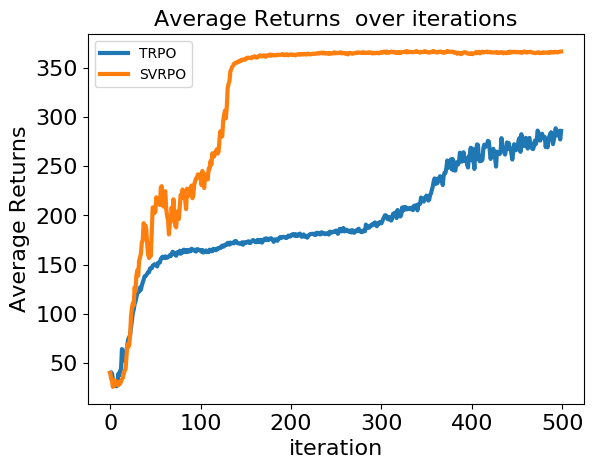}
\caption{\label{sfig:swimmer} Swimmer.}
\end{subfigure}
\begin{subfigure}[b]{0.3 \textwidth}
\includegraphics[width=\textwidth]{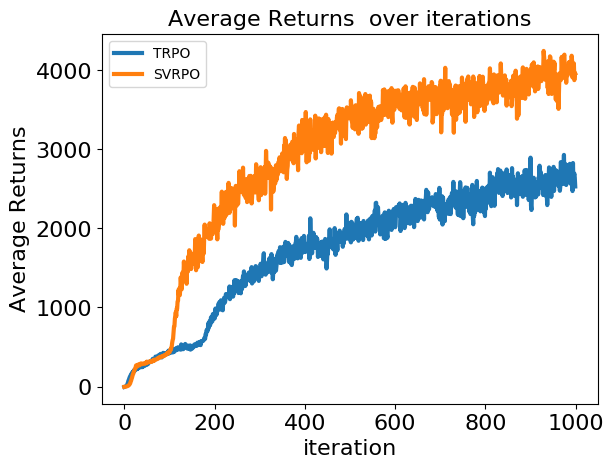}
\caption{\label{sfig:walker} Walker}
\end{subfigure}
\begin{subfigure}[b]{0.3 \textwidth}
\includegraphics[width=\textwidth]{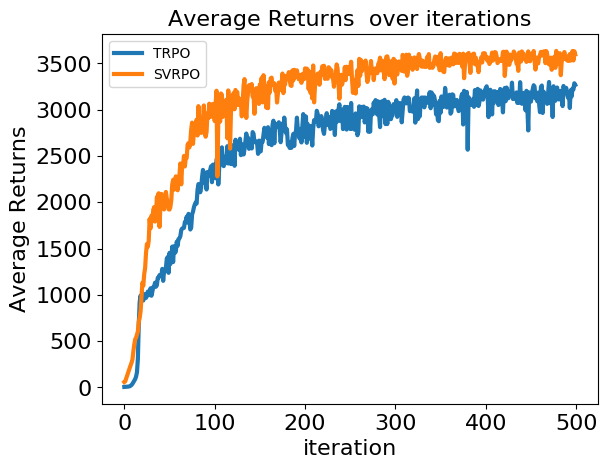}
\caption{\label{sfig:hopper} Hopper}
\end{subfigure} 
\\
\begin{subfigure}[b]{0.3 \textwidth}
\includegraphics[width=\textwidth]{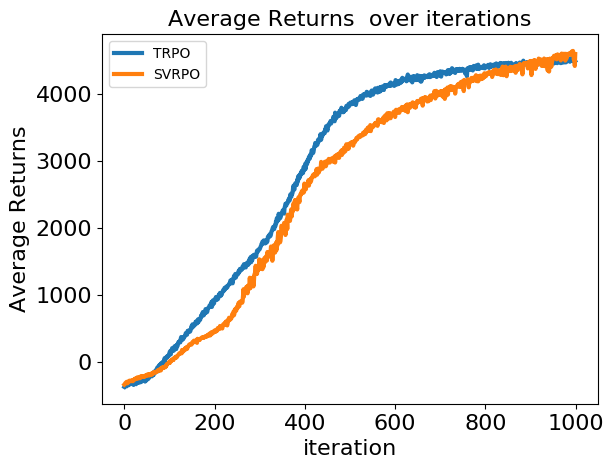}
\caption{\label{sfig:cheetah} Half-Cheetah}
\end{subfigure}
\begin{subfigure}[b]{0.3 \textwidth}
\includegraphics[width=\textwidth]{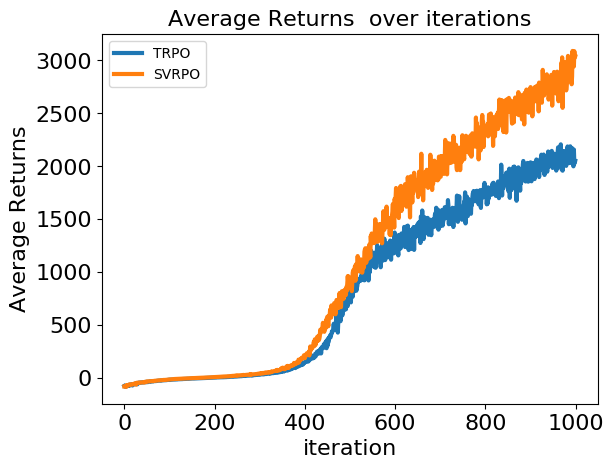}
\caption{\label{sfig:ant} Ant}
\end{subfigure}
\begin{subfigure}[b]{0.3 \textwidth}
\includegraphics[width=\textwidth]{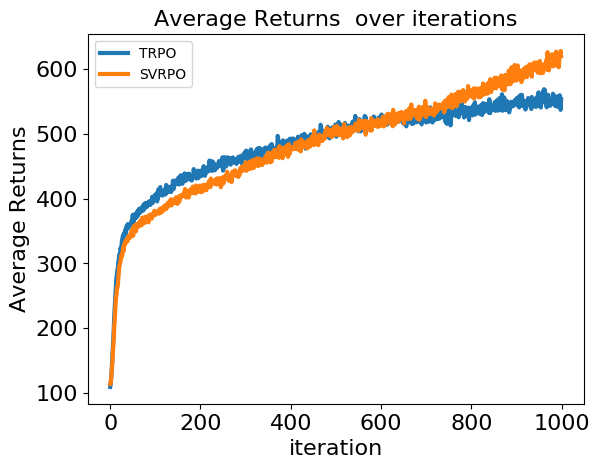}
\caption{\label{sfig:humanoid} Humanoid}
\end{subfigure}
\caption{Performance Comparison of SVRPO and TRPO for six Mujoco Control Tasks.} \label{fig:mujoco}
\end{figure*}

\begin{table}
\centering
\begin{tabular}{l|r}
& TRPO and SVRPO \\\hline
Num steps per Iter. & 50,000 \\
Discount($\lambda$) & 0.995 \\
Horizon             & 1,000 \\
Step Size($\delta$) & 0.01 \\
\end{tabular}
\caption{\label{tab:trpo} Parameter settings for TRPO and SVRPO}
\end{table}

\paragraph{Experimental Settings}
All the experiments are doing in OpenAI rllab development toolkit \cite{duan:benchmark}.  
For a fair comparison, we set the common hyper-parameters of TRPO and SVRPO to be the same according to the continuous control benchmark \cite{duan:benchmark}; see Table \ref{tab:trpo}. 
We use a Gaussian Policy with a diagonal covariance matrix, whose mean is parameterized by  a multi-layer perceptron (MLP) with $tanh$ activation function. 
For Ant and Humanoid, 
we use a Gaussian MLP with three hidden layers of size (64,64,64) in the policy networks; for the other tasks, 
we use  two hidden layers of size (64,64). 
To trade-off the accuracy and computations of estimation of Fisher information matrix, we set the subsample ratio $\nu = 10\%$. 
We search the parameter $J$ (the number of inner loop iterations) in the range of $[10, 100]$ along with different mini-batch sizes $m$ tuned as either $1000$ or $5000$ for most tasks. A large value of $m$ would introduce more computational cost, 
while smaller $m$ may not scan all the samples in one pass to get enough information.  A useful intuition is to set $m J \approx N$ or $m J \approx 2N$, so that we effectively go through one or two passes of the trajectory set.

\paragraph{Performance and Sample Efficiency}
Fig.\ref{fig:mujoco} plots the learning curves of SVRPO and TRPO for various Mujoco tasks. 
Since each iteration of SVRPO and TRPO use the same number of samples, 
the $x$-axes are also proportional to the number of samples we use in each algorithm.  
We can see that for Swimmer (\ref{sfig:swimmer}), SVRPO converges much faster and achieves much higher return than TRPO, taking only $\leq 40\%$ samples to achieve similar performance as the final results of TRPO. 
Similarly, for Hopper (\ref{sfig:hopper}), less than 200 iterations of SVRPO can achieve similar results as that of TRPO in 500 iterations, and SVRPO achieves much higher final return than TRPO. 
For Walker (\ref{sfig:walker}), our SVRPO only needs  $\leq 40\%$ of samples of TRPO to have the similar return, and achieves significant better performance with return as high as about 4000 compared to TRPO's about 2900. 
In Ant (\ref{sfig:ant}), our SVRPO also achieves sample efficiency and significant better return at about 3000 compared to TRPO's about 2200 for 1000 iterations. 
For Humanoid (\ref{sfig:humanoid}), SVRPO is able to achieve similar or slightly better return in 1000 iterations. 
In Half-Cheetah (\ref{sfig:cheetah}), our SVRPO has similar final return as TRPO in 1000 iterations.

\paragraph{Implicit Exploration with parameter noise}
SVRPO adopts sub-sampled Fisher information matrix and estimate the variance reduction policy gradient with mini-batch; these design choices introduce small, controlled 
noise which randomly moves smoothly around the promising regions in the parameter space, without introducing large and wild jumps, this provides an implicit and 'safe' 
exploration mechanism. 
This exploration is beneficial to sample efficiency as it encourage to learn fast within a 
wider range of parameter space especially in the early stage when the network parameters 
are not well fine-trained. 
With wider range of parameters to consider, it will end up with high probability to find a 
better parameter region for the policy to be optimal.
For example, as Hopper (\ref{sfig:hopper}) has more exploration with injected noise, 
our SVRPO learns much faster in the early stage with only about $30\%$ samples 
to achieve the similar return as TRPO. 
We also have similar observations for control tasks such as Swimmer, Walker, etc.

\begin{figure*}
\centering
\begin{subfigure}[b]{0.3 \textwidth}
\includegraphics[width=\textwidth]{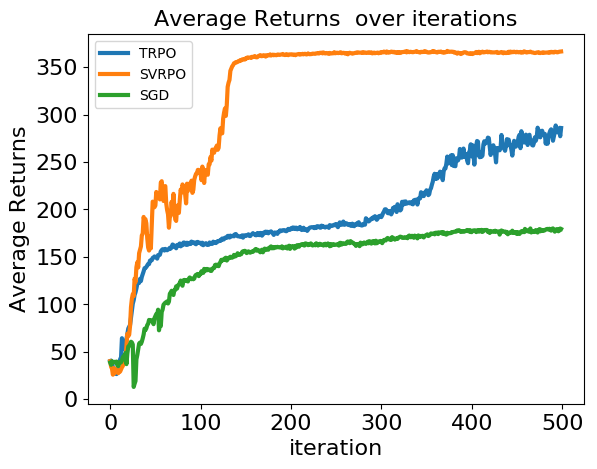}
\caption{\label{sfig:swimmer_sgd} Swimmer.}
\end{subfigure}
\begin{subfigure}[b]{0.3 \textwidth}
\includegraphics[width=\textwidth]{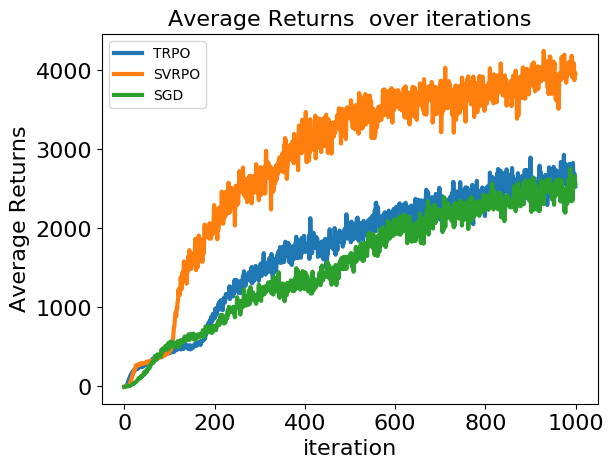}
\caption{\label{sfig:walker_sgd} Walker}
\end{subfigure}
\caption{Performance Advantage of Variance Reduction Policy Gradient (SVRG vs SGD in our SVRPO) with baseline TRPO for Swimmer and Walker.} \label{fig:sgd}
\end{figure*}

\begin{figure*}
\centering
\begin{subfigure}[b]{0.3 \textwidth}
\includegraphics[width=\textwidth]{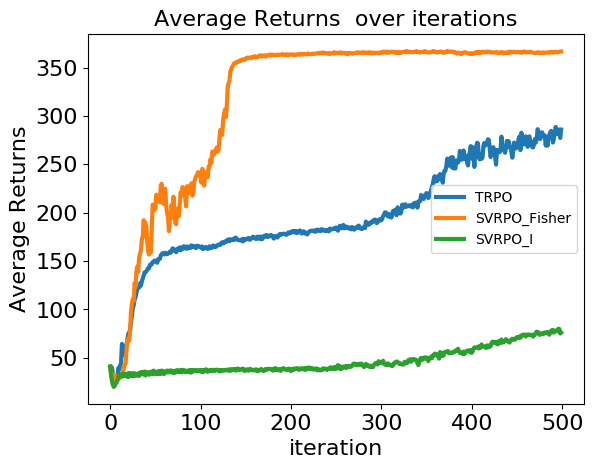}
\caption{\label{sfig:swimmer_fisher} Swimmer.}
\end{subfigure}
\begin{subfigure}[b]{0.3 \textwidth}
\includegraphics[width=\textwidth]{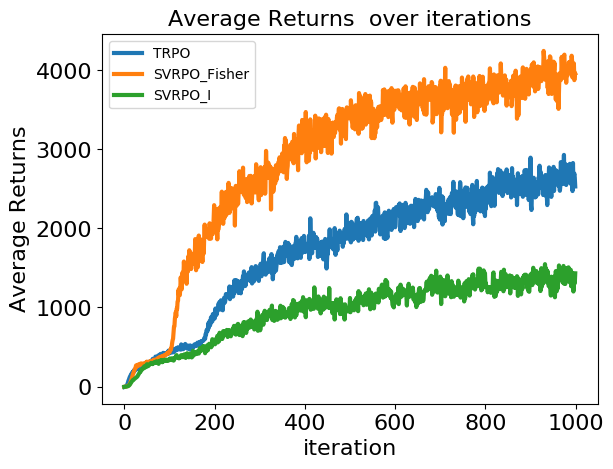}
\caption{\label{sfig:walker_fisher} Walker.}
\end{subfigure}
\caption{Fisher information matrix's accelerated convergence rates on our SVRPO with baseline TRPO for Swimmer and Walker.} \label{fig:fisher}
\end{figure*}

\subsection{Understanding SVRPO}
We take investigation on our SVRPO algorithm in term of the performance improvement from control variates for the variance reduction and convergence acceleration impacts from sub-sampled Fisher information matrix. Here, all the experimental settings are the same as subsection (\ref{sec:mujoco}).

In Fig.(\ref{fig:sgd}), it is clear that stochastic variance reduction gradient (Eq.\ref{eq:spg}) is the major weapon to achieve our significant better results than TRPO. Here, SGD denotes the algorithm that we replace with the policy gradient (Eq.\ref{eq:minipg}) instead of SVRG estimation (Eq.\ref{eq:spg}). It could be regarded as a stronger version of \textbf{REINFORCE} with additional line search.
For a clear illustration, the update of SVRG in SVRPO is,
\begin{align*}
w \gets w + \eta H^{-1} \hat{\nabla}_{sv} U(w)
\end{align*}
and SGD in SVRPO is,
\begin{align*}
w \gets w + \eta \hat{\nabla} U(w)
\end{align*}
We also compare SVRPO (SVRG and SGD) with TRPO,
\begin{align*}
w \gets w + \eta H^{-1} \hat{\nabla} U(w)
\end{align*}

As the estimation of policy gradients introduce very high variance due to long horizon, mini-batch estimation noise, unknown environment dynamics and so on, it is necessary to reduce the variance for the gradient estimation to achieve better policies.
As you can see for Swimmer(Fig.\ref{sfig:swimmer_sgd}) with vanilla policy gradient (SGD), it stuck in bad local policies and difficult to improve the policies due to high variance inaccurate estimation. Small controlled noise might be beneficial to exploration in the parameter space, however highly wild variance would eventually hurt the performance due to inaccurate noisy estimation as illustrate in both Swimmer(Fig.\ref{sfig:swimmer_sgd}) and Walker(Fig.\ref{sfig:walker_sgd}).

Another major weapon is the Fisher information matrix first introduced in natural gradient and then successfully adopted in TRPO. Curvature information is very helpful to accelerate the convergence rates as illustrated in Fig.(\ref{fig:fisher}). Without adopting Fishing information (with $I$ replacing $H$ in Eq.(\ref{eq:newton})), the performance gets stuck in very bad policies and it is difficult for the agent to move forward and eventually needs much more number of samples (iterations) to achieve convergence as showed in Fig.(\ref{sfig:swimmer_fisher}) and Fig.(\ref{sfig:walker_fisher}) for Swimmer and Walker.

Consequently, our SVRPO combines both variance reduction technique and Fisher information matrix to further achieve significant better returns with sample efficiency for many Mujoco tasks such as Swimmer, Walker, etc.



\vspace {0.5 cm}
\section{Related Work}
In reinforcement learning\cite{sutton:rl}, policy search (or policy optimization) 
is to find the optimal policy parameterized with linear function approximation or highly non-linear functions such as neural networks. 
It has wide applications in robotic learning \cite{peters:robotics, peters:motor} with continuous action space and high-dimensional state space, 
for example from robotics locomotion \cite{stone:pg, seung:pgbiped, levine:robot}
to manipulation \cite{deisenroth:ps,levine:e2e}, 
and robust policy search for safe vehicle navigation \cite{kobilarov:rps}, 
model based policy search for robot control \cite{deisenroth:robot, deisenroth:pilco}, 
multi-robot coordination policy search \cite{kaelbling:mps} and so on.
Our work is also inspired by the stochastic variance reduction for policy evaluation \cite{xiao:svrg}.

Optimization methods \cite{nocedal:opt, nestrov:convex, bertsekas:opt} 
play a key role in the policy search, especially for nonlinear policies in continuous high-dimensional parameter space. For example, the well-known \textbf{REINFORCE} \cite{williams:vpg} is simply a (stochastic) gradient descent method. To accelerate the convergence rates, Fisher information is adopted in Natural Gradient \cite{kakade:ng, peters:nac} and TRPO \cite{schulman:trpo}. Stochastic Variance Reduction 
\cite{tong:svrg} is proposed under the mechanics of control variates \cite{owen:control} to accelerate the convergence of SGD by dramatic variance reduction. Recently, second order statistics and stochastic curvature information are adopted \cite{nocedal:hessian, nocedal:subsample, jordan:lbfgs, singer:sqn} to improve the convergence while achieving 
the good trade-off between computations and accuracy for the large-scale machine learning problems. The stochastic and approximated curvature information 
is also useful to further accelerate the variance reduction methods \cite{zhang:vrp, erdogdu:svrg}. 

\vspace {0.5 cm}
\section{Conclusion and Discussion}
Unbounded and high variance in policy gradient estimation is a major concern and often lead to poor performance in the model-free reinforcement learning settings. 
In this paper, we developed a trust region stochastic variance reduction method for policy gradient estimation to optimize the policies with application to robotic continuous control problems. 
We applied the variance reduced policy gradient estimation with a control variate into a trust region Newton-CG optimization framework. 
Our method also introduces small controlled noise from stochastic mini-batch estimation to encourage 
exploration in the parameter space. 
Furthermore, multiple mini-batch updates are able to make more efficient usage of the 
information in the trajectories and thus introduces possible benefits to obtain better sample efficiency. 
Systematic experiments show that our SVRPO achieves better performance with 
improved sample efficiency in Mujoco tasks such as 
Swimmer, Walker, Hopper and Ant compared to TRPO.  

For the future work, our stochastic variance reduction method opens door for the further development of scalable 
and highly efficient methods beyond the stochastic gradient descent for policy optimization.
As a stochastic mini-batch algorithm, our SVRPO is able to adopt to large-scale reinforcement learning 
and robotics problems with very long horizons and large number of samples. 
The sub-sampled estimation of the Fisher information matrix is able to further trade-off the computations and 
accuracy, which is especially useful in large-scale problems.
Secondly, it is also promising to extend our work into robotic learning problems from 
large state space such as virtual observations to complex control policies similar to
\cite{levine:e2e}.

\addtolength{\textheight}{-12cm}   



\bibliographystyle{IEEEtran}
\bibliography{root}

\begin{thebibliography}{10}
\providecommand{\url}[1]{#1}
\csname url@rmstyle\endcsname
\providecommand{\newblock}{\relax}
\providecommand{\bibinfo}[2]{#2}
\providecommand\BIBentrySTDinterwordspacing{\spaceskip=0pt\relax}
\providecommand\BIBentryALTinterwordstretchfactor{4}
\providecommand\BIBentryALTinterwordspacing{\spaceskip=\fontdimen2\font plus
\BIBentryALTinterwordstretchfactor\fontdimen3\font minus
  \fontdimen4\font\relax}
\providecommand\BIBforeignlanguage[2]{{%
\expandafter\ifx\csname l@#1\endcsname\relax
\typeout{** WARNING: IEEEtran.bst: No hyphenation pattern has been}%
\typeout{** loaded for the language `#1'. Using the pattern for}%
\typeout{** the default language instead.}%
\else
\language=\csname l@#1\endcsname
\fi
#2}}

\bibitem{tong:svrg}
R.~Johnson and T.~Zhang, ``Accelerating stochastic gradient descent using
  predictive variance reduction,'' in \emph{Proceedings of the 26th
  International Conference on Neural Information Processing Systems (NIPS)},
  2013.

\bibitem{nocedal:opt}
J.~Nocedal and S.~J. Wright, \emph{Numerical Optimization}.\hskip 1em plus
  0.5em minus 0.4em\relax New York, NY, USA: Springer, 2006.

\bibitem{schulman:trpo}
J.~Schulman, S.~Levine, P.~Moritz, M.~I. Jordan, and P.~Abbeel, ``Trust region
  policy optimization,'' in \emph{Proceedings of the 32nd International
  Conference on Machine Learning (ICML)}, 2015.

\bibitem{sutton:pg}
R.~S. Sutton, D.~McAllester, S.~Singh, and Y.~Mansour, ``Policy gradient
  methods for reinforcement learning with function approximation,'' in
  \emph{NIPS'99 Proceedings of the 12th International Conference on Neural
  Information Processing Systems}, 1999, pp. 1057--1063.

\bibitem{williams:vpg}
R.~J. Williams, ``Simple statistical gradient-following algorithms for
  connectionist reinforcement learning,'' \emph{Machine Learning}, vol.~8, p.
  229–256, 1992.

\bibitem{kakade:ng}
S.~Kakade, ``A natural policy gradient,'' in \emph{Proceedings of the 14th
  International Conference on Neural Information Processing Systems
  (NIPS-01)}.\hskip 1em plus 0.5em minus 0.4em\relax MIT Press, 2001, pp.
  1531--1538.

\bibitem{mnih:dqn}
V.~Mnih, K.~Kavukcuoglu, D.~Silver, A.~A. Rusu, J.~Veness, M.~G. Bellemare,
  A.~Graves, M.~Riedmiller, A.~K. Fidjeland, G.~Ostrovski, S.~Petersen,
  C.~Beattie, A.~Sadik, I.~Antonoglou, H.~King, D.~Kumaran, D.~Wierstra,
  S.~Legg, and D.~Hassabis, ``Human-level control through deep reinforcement
  learning,'' \emph{Nature}, vol. 518, pp. 529--533, 2015.

\bibitem{todorov:mujoco}
E.~Todorov, T.~Erez, and Y.~Tassa, ``Mujoco: A physics engine for model-based
  control,'' \emph{2012 IEEE/RSJ International Conference on Intelligent Robots
  and Systems}, pp. 5026--5033, 2012.

\bibitem{nocedal:hessian}
R.~H. Byrd, G.~M. Chin, W.~Neveittand, and J.~Nocedal, ``On the use of
  stochastic hessian information in optimization methods for machine
  learning,'' \emph{SIAM Journal on Optimization (SIOPT)}, vol.~21, p.
  977–995, 2011.

\bibitem{nocedal:subsample}
\BIBentryALTinterwordspacing
R.~Bollapragada, R.~Byrd, and J.~Nocedal, ``Exact and inexact subsampled newton
  methods for optimization,'' \emph{arXiv}, 2016. [Online]. Available:
  \url{https://arxiv.org/pdf/1609.08502.pdf}
\BIBentrySTDinterwordspacing

\bibitem{erdogdu:svrg}
R.~Kolte, M.~Erdogdu, and A.~Ozgür, ``Accelerating svrg via second-order
  information,'' \emph{In NIPS Workshop on Optimization for Machine Learning},
  2015.

\bibitem{zhang:vrp}
\BIBentryALTinterwordspacing
J.~Wang and T.~Zhang, ``Improved optimization of finite sums with minibatch
  stochastic variance reduced proximal iterations,'' \emph{arXiv}, 2017.
  [Online]. Available: \url{https://arxiv.org/pdf/1706.07001.pdf}
\BIBentrySTDinterwordspacing

\bibitem{duan:benchmark}
Y.~Duan, X.~Chen, R.~Houthooft, J.~Schulman, and P.~Abbeel, ``Benchmarking deep
  reinforcement learning for continuous control,'' in \emph{Proceedings of the
  33rd International Conference on Machine Learning (ICML)}, 2016.

\bibitem{sutton:rl}
R.~S. Sutton and A.~G. Barto, \emph{Introduction to Reinforcement
  Learning}.\hskip 1em plus 0.5em minus 0.4em\relax Cambridge, MA, USA: MIT
  Press, 1998.

\bibitem{peters:robotics}
M.~P. Deisenroth, G.~Neumann, and J.~Peters, ``A survey on policy search for
  robotics,'' \emph{Foundations and Trends in Robotics}, vol.~2, pp. 1--142,
  2013.

\bibitem{peters:motor}
J.~Peters and S.~Schaal, ``Reinforcement learning of motor skills with policy
  gradients,'' \emph{Neural Networks}, vol.~21, pp. 682--697, 2008.

\bibitem{stone:pg}
N.~Kohl and P.~Stone, ``Policy gradient reinforcement learning for fast
  quadrupedal locomotion,'' in \emph{Proceedings of the IEEE International
  Conference on Robotics and Automation (ICRA 2004)}, 2004.

\bibitem{seung:pgbiped}
R.~Tedrake, T.~W. Zhang, and H.~S. Seung, ``Stochastic policy gradient
  reinforcement learning on a simple 3d biped,'' in \emph{IEEE/RSJ
  International Conference on Intelligent Robots and Systems (IROS 2004)},
  2004.

\bibitem{levine:robot}
S.~Levine, N.~Wagener, and P.~Abbeel, ``Learning contact-rich manipulation
  skills with guided policy search,'' in \emph{Proceedings of the IEEE
  International Conference on Robotics and Automation (ICRA 2015)}, 2015.

\bibitem{deisenroth:ps}
M.~P. Deisenroth, P.~Englert, J.~Peters, and D.~Fox, ``Multi-task policy search
  for robotics,'' in \emph{Proceedings of the IEEE International Conference on
  Robotics and Automation (ICRA 2014)}, 2014.

\bibitem{levine:e2e}
S.~Levine, C.~Finn, T.~Darrell, and P.~Abbeel, ``End-to-end training of deep
  visuomotor policies,'' \emph{The Journal of Machine Learning Research},
  vol.~17, pp. 1334--1373, 2016.

\bibitem{kobilarov:rps}
M.~Sheckells, G.~Garimella, and M.~Kobilarov, ``Robust policy search with
  applications to safe vehicle navigation,'' in \emph{Proceedings of the IEEE
  International Conference on Robotics and Automation (ICRA 2017)}, 2017, pp.
  2343--2349.

\bibitem{deisenroth:robot}
B.~Bischoff, D.~Nguyen-Tuong, H.~van Hoof, A.~McHutchon, C.~Rasmussen,
  A.~Knoll, J.~Peters, and M.~Deisenroth, ``Policy search for learning robot
  control using sparse data,'' in \emph{Proceedings of the IEEE International
  Conference on Robotics and Automation (ICRA 2014)}, 2014, pp. 3882--3887.

\bibitem{deisenroth:pilco}
M.~P. Deisenroth and C.~E. Rasmussen, ``Pilco: A model-based and data-efficient
  approach to policy search,'' in \emph{Proceedings of the 28th International
  Conference on Machine Learning (ICML 2011)}, 2011.

\bibitem{kaelbling:mps}
C.~Amato, G.~Konidaris, A.~Anders, G.~Cruz, J.~P. How, and L.~P. Kaelbling,
  ``Policy search for multi-robot coordination under uncertainty,'' \emph{The
  International Journal of Robotics Research}, vol.~35, 2017.

\bibitem{xiao:svrg}
S.~S. Du, J.~Chen, L.~Li, L.~Xiao, and D.~Zhou, ``Stochastic variance reduction
  methods for policy evaluation,'' in \emph{Proceedings of the International
  Conference on Machine Learning(ICML)}, 2017.

\bibitem{nestrov:convex}
Y.~Nesterov, \emph{Introductory Lectures on Convex Optimization}.\hskip 1em
  plus 0.5em minus 0.4em\relax Springer, 2014.

\bibitem{bertsekas:opt}
D.~P. Bertsekas, \emph{Nonlinear Programming}.\hskip 1em plus 0.5em minus
  0.4em\relax Belmont, MA, USA: Athena Scientific, 1999.

\bibitem{peters:nac}
J.~Peters and S.~Schaal, ``Natural actor-critic,'' \emph{Neurocomputing},
  vol.~71, pp. 1180--1190, 2008.

\bibitem{owen:control}
A.~Owen and Y.~Zhou, ``Safe and effective importance sampling,'' \emph{Journal
  of the American Statistical Association}, vol.~95, pp. 135--143, 2000.

\bibitem{jordan:lbfgs}
R.~N. P.~Moritz and M.~Jordan, ``A linearly-convergent stochastic l-bfgs
  algorithm,'' in \emph{Proceedings of the 19th International Conference on
  Artificial Intelligence and Statistics (AISTATS)}, 2016, pp. 249--258.

\bibitem{singer:sqn}
R.~H. Byrd, S.~L. Hansen, J.~Nocedal, and Y.~Singer, ``A stochastic
  quasi-newton method for large-scale optimization,'' \emph{SIAM Journal on
  Optimization}, vol.~26, pp. 1008--1031, 2016.

\end{thebibliography}

\end{document}